\documentclass{article}

\usepackage[T1]{fontenc}

\usepackage{url}            
\usepackage{amsfonts}       
\usepackage{nicefrac}       

\usepackage{amsmath,amssymb,amsthm,amsfonts}
\usepackage{mathrsfs}
\usepackage{xspace}
\usepackage{algorithm}
\usepackage{algorithmic}
\usepackage{bm}
\usepackage{color}
\usepackage{multirow}
\usepackage{hhline}
\usepackage{ulem}
\usepackage{xcolor}

\usepackage{lipsum}
\usepackage{float}
\usepackage{zref-xr}
\zxrsetup{toltxlabel=true, tozreflabel=false}

\renewcommand{\algorithmicrequire}{\textbf{Input:}}
\renewcommand{\algorithmicensure}{\textbf{Output:}}

\newtheoremstyle{case}{}{}{}{}{}{:}{ }{}
\theoremstyle{case}
\newtheorem{case}{Case}

\newcommand\M{\mathcal{M}}

\newcommand\oneshot{\mathcal{M}^{\mathrm{os}}}

\newcommand{\ep}[1]{\ifthenelse{\equal{#1}{1}}{\mathrm{e}}{\mathrm{e}^{#1}}}

\def\prob{\mathbb{P}}
\let\P\prob
\def\lap{\operatorname{Lap}}

\let\e\varepsilon
\let\eps\varepsilon
\renewcommand\epsilon{\varepsilon}

\let\emph\textit

\newcommand{\remove}[1]{}
\let\mathbbm\textbf

\newtheorem{othertheorem}{othertheorem}[section]

\theoremstyle{lemma}
\newtheorem{lemma}[othertheorem]{Lemma}
\theoremstyle{corollary}

\theoremstyle{proposition}

\theoremstyle{theorem}
\newtheorem{theorem}[othertheorem]{Theorem}

\theoremstyle{definition}
\newtheorem{definition}[othertheorem]{Definition}
\theoremstyle{remark}

\theoremstyle{assumption}

\allowdisplaybreaks
\numberwithin{equation}{section}

\usepackage{hyperref}

\usepackage[accepted]{icml2021}

\icmltitlerunning{Oneshot Differentially Private Top-$k$ Selection}

\begin{document}

\twocolumn[
\icmltitle{Oneshot Differentially Private Top-$k$ Selection}



\icmlsetsymbol{equal}{*}

\begin{icmlauthorlist}
\icmlauthor{Gang Qiao}{to}
\icmlauthor{Weijie J.~Su}{goo}
\icmlauthor{Li Zhang}{ed}
\end{icmlauthorlist}

\icmlaffiliation{to}{Department of Statistics, University of Michigan, Ann Arbor, MI, USA.}
\icmlaffiliation{goo}{The Wharton School, University of Pennsylvania, Philadelphia, PA, USA.}
\icmlaffiliation{ed}{Google Research, Mountain View, CA, USA}

\icmlcorrespondingauthor{}{\texttt{qiaogang@umich.edu, suw@wharton.upenn.edu, liqzhang@google.com}}

\icmlkeywords{Machine Learning, ICML}

\vskip 0.3in
]



\printAffiliationsAndNotice{}  

\begin{abstract}
Being able to efficiently and accurately select the top-$k$ elements with differential privacy is an integral component of various private data analysis tasks. In this paper, we present the oneshot Laplace mechanism, which generalizes the well-known Report Noisy Max~\cite{dwork2014algorithmic} mechanism to reporting noisy top-$k$ elements. We show that the oneshot Laplace mechanism with a noise level of $\widetilde{O}(\sqrt{k}/\eps)$ is approximately differentially private. Compared to the previous peeling approach of running Report Noisy Max $k$ times, the oneshot Laplace mechanism only adds noises and computes the top $k$ elements once, hence much more efficient for large $k$. In addition, our proof of privacy relies on a novel coupling technique that bypasses the use of composition theorems. Finally, we present a novel application of efficient top-$k$ selection in the classical problem of ranking from pairwise comparisons.
\end{abstract}


\section{Introduction}
\label{sec:introduction}

Modern statistical analyses have increasingly relied on sensitive data from individuals and, accordingly, there is a growing recognition that privacy constraints should be incorporated into consideration in data analysis. In response, a mathematically rigorous framework called \emph{differential privacy} \citep{DworkKMMN06,DworkMNS06} was introduced for privacy-preserving data analysis. Roughly speaking, a differentially private procedure ensures that the released information is not influenced significantly by any individual record in the dataset. As a consequence, the privacy of the individuals will not be revealed based on the released information.

This paper is concerned with the top-$k$ problem, one of the most important primitives in differential privacy: reporting $k$ items with (approximately) the maximum values among $m$ given values. The problem of \textit{privately} reporting the $k$ largest elements is an essential building block in many machine learning tasks and has gained continued popularity in the literature
\citep{recom09,friedman2010data,banerjee2012price,mcsherry2013differential,shen2014privacy,qin2016heavy,bafna2017price,steinke2017tight,DworkDPFDRC18,durfee2019practical}. The common peeling solution~\citet{hardt2013beyond} and \citet{DworkDPFDRC18} is by iteratively applying the Report Noisy Max algorithm and then resorting to the composition theorem for computing the privacy loss. In general, this results in the noise level of $O(k/\eps)$ for $\eps$ pure privacy and $\widetilde{O}(\sqrt{k}/\eps)$\footnote{Throughout the paper, we use $\widetilde{O}$ to hide dependence on logarithmic factors.} for $(\eps,\delta)$ privacy loss. While the peeling algorithm has good privacy guarantee, it requires to run Report Noisy Max $k$ times, hence incurring a high computational cost for large $k$. 

In this paper, we show that by adapting Report Noisy Max to reporting noisy top-$k$ items, we can still achieve comparable utility but with a much more efficient procedure as we only need to run the selection once. We call the resulted algorithm \emph{the oneshot Laplace mechanism}. More precisely, in the oneshot Laplace mechanism, we add the Laplace noise to each count and then report the \emph{set} of items with the top-$k$ noisy counts. In this paper, we show that the oneshot Laplace mechanism can achieve utility comparable to those obtained from the peeling procedure~(see Theorems~\ref{thm:simple} and \ref{thm:main} for the precise statements).

It is relatively straightforward to show that by adding Laplace noise of level $k/\eps$, the mechanism is $\eps$ purely differentially private~(Theorem~\ref{thm:simple}).\footnote{This is probably a folklore but since we could not find a reference, we include its proof for completeness.} However, it turns out to be much more challenging to show that with $\widetilde{O}(\sqrt{k}/\eps)$ noise, it is $(\eps,\delta)$-differentially private. Indeed, the proof of this fact is the main contribution of our paper~(Theorem~\ref{thm:main}).

Our proof directly bounds the privacy loss without the help of the composition theorems. The difficulty of this approach is in untangling the complex distribution \textit{dependence} of the $k$ selected items, as opposed to the \textit{conditional dependence} in other existing mechanisms, which allows us to use the advanced composition theorem (\citet{boosting}, see also \citet{KOV}). To deal with this difficulty, we introduce a novel theoretical technique that, in effect, reduces the oneshot problem to a multinomial distribution problem to bypass the use of composition theorems. To shed light on our proof, consider the case of many similar or equal values. This is the case where the true top-$k$ set can be extremely sensitive to the change of input values. In order to privately report the top-$k$ set in this case, we add independent Laplace noises centered at zero to these values, which yields an approximately equal chance that the noisy values of two adjacent inputs will ``go up'' or ``go down'',  leading to the cancellation of certain first-order terms in the (logarithms of) the probabilities of events and hence a tight control between their ratio.

Circumventing composition theorems, however, may have its advantage. Since it relies on the direct analysis so may avoid the slackness introduced by the generic composition theorems. Indeed, there have been recent work on exploring the special properties of the privacy mechanisms to improve upon the generic composition theorem~\cite{abadi2016deep,bun2016concentrated,dong}.

One closely related previous work is the oneshot Gumbel mechanism proposed in \citet{durfee2019practical}. In that paper, the authors show that adding Gumbel noise and reporting the top-$k$ items is \emph{equivalent} to the peeling procedure of the exponential mechanism for reporting the maximum item~\cite{dwork2014algorithmic}. This elegant connection immediately provides privacy guarantees through the well understood composition of exponential mechanisms and can benefit from any improvement of the composition property~\cite{dong2019optimal}. In addition, their mechanism reports the noisy rank of the top-$k$ selections. However, it is important to note that their analysis goes through the composition theorem, whereas our paper takes an entirely different angle by employing a composition-free analysis. The comparison between these two approaches, both in theory and in practice, remains an interesting future work.

\subsection{Preliminaries}
\label{sec:prel-diff-priv}

Before continuing, we pause to revisit some basic concepts in differential privacy.

\begin{definition}
Data sets $D, D'$ are said to be \emph{neighbors}, or \emph{adjacent}, if one is obtained by removing or adding a single data
item.
\end{definition}
Differential privacy, sometimes called \emph{pure differential privacy} now, was first defined and constructed in~\citet{DworkMNS06}.  The
relaxation of pure differential privacy defined next is sometimes referred to as \emph{approximate
differential privacy} or \emph{$(\e,\delta)$-differential privacy}.
\begin{definition}[Differential privacy~\citep{DworkMNS06}] 
\label{def:dp}
A randomized mechanism $\M$ is $(\e,\delta)$-differentially private if
for all adjacent $D,D'$, and for any subset of possible outputs $S$: $\prob(\M(D) \in S)\leq
\ep{\e} \prob(\M(D') \in S) + \delta$. 
Pure differential privacy is the special case of approximate differential privacy in which~$\delta=0$.
\end{definition}
In differential privacy problems, privacy law protects individually identifiable data, and the parameters $\e$ and $\delta$  in the definition above measure the degree of privacy protected. Let $f=f(D)$ be a statistic on a database $D$. In any randomized $(\e,\delta)$-differentially private mechanism $\M$, the perturbed response $f(D)+Z$ is reported instead of the true answer $f(D)$, where $Z$ is the random noise to guarantee indistinguishability between two datasets. The sensitivity of the statistic, or query function $f$, is the largest change in its output when we change a single data item and is defined below.

\begin{definition}[Sensitivity]
\label{def:sensitivity2}
Let $f=(f_1, \cdots, f_m)$ be $m$ real valued functions that take a database as input. The sensitivity of $f$, denoted as $s$, is defined as
\[
s_f = \max_{D,D'} \max_{1 \le i \le m} |f_i(D)-f_i(D')| \,,
\]
where the maximum is taken over any adjacent databases $D$ and $D'$.
\end{definition}

In the \emph{Laplace Mechanism}, the output $f(D)$ is perturbed with noise generated from the Laplace distribution $\lap(\lambda)$ with probability density function: $f_{\lap(\lambda)}(z)=\frac{1}{2\lambda} \ep{-|z|/\lambda}$, where the scale $\lambda$ should be calibrated to the sensitivity of the statistics $f$.



\section{The Oneshot Laplace Mechanism}
\label{sec:oneshot-mechanism}

In this section, we introduce the oneshot Laplace mechanism in full detail, along with its privacy guarantees. Consider the problem of privately reporting the minimum  $k$ locations of $m$ values $x_1, \ldots, x_m$ and their estimated values.  Here two input values $(x_1, \ldots, x_m)$ and $(x'_1, \ldots, x'_m)$ are called \emph{adjacent} if $\|x-x'\|_\infty\leq 1$, i.e., $|x_i-x'_i|\leq 1$ for all $1\leq i\leq m$. In this definition, $x$ can be considered as the counts of each of $m$-attributes of the population in some database $D$ and, therefore, changing any individual in $D$ may in the worst case change each count $x_i$ by $1$.

As a special case when $k=1$, the solution relies on the Report Noisy Min algorithm~\citep{dwork2014algorithmic,DworkDPFDRC18}, which takes as input a function $f$, database $D$, and privacy parameter $\e$, and outputs the index of the minimum element and its estimated value. The Report Noisy Min algorithm adds independently sampled $\lap(2s_f/\e)$ noise to each element of $f(D)$ and reports the index $i^\ast$ of the minimum noisy count. The algorithm further reports its estimated value by adding noise freshly sampled from $\lap(2s_f/\e)$ to $f_{i^\ast}(D)$. In ~\citet{dwork2014algorithmic,DworkDPFDRC18}, the Report Noisy Min algorithm is proved to be $(\e,0)$-differentially private. Notably, in order to avoid violation of differential privacy, we shall not report the minimum noisy element as its estimated value. Hence, we need to add fresh random noise to  $f_{i^\ast}(D)$ in the last step.

To efficiently solve the top-$k$ problem where $k$ can be larger than 1, we introduce the oneshot Laplace mechanism $\oneshot$, which is one of our main contributions in this paper. In $\oneshot$, we add noise $\lap(\lambda)$ once to each value and report the indices and approximations of the minimum $k$ noisy values (Algorithm~\ref{algo:oneshot}).

\begin{algorithm}[ht]
\caption{The Oneshot Laplace Mechanism $\oneshot$ for Privately Reporting Minimum $k$ Elements}\label{algo:oneshot}
\begin{algorithmic}[1]
\REQUIRE database $D$, functions $f=(f_1,\cdots,f_m)$ with sensitivity $s_f$, parameter $k$, and the noise scale $\lambda$
\ENSURE indices $i_1, \ldots, i_k$ and approximations to $f_{i_1}(D), \cdots, f_{i_k}(D)$
\FOR {$i=1$ to $m$}
  \STATE set $y_i = f_i(D)+g_i$ where $g_i$ is sampled i.i.d. from $\lap(\lambda)$
\ENDFOR
\STATE sort $y_1,\ldots, y_m$ from low to high, $y_{i_1} \le y_{i_2} \le \cdots \le y_{i_m}$
\STATE return the set $\{i_1, \ldots, i_k\}$ and $f_{i_j}(D)+g'_{i_j}$, where $1 \le j \le k$ and $g'_{i_j}$ are fresh independent random noise sampled from $\lap(\lambda)$
\end{algorithmic}
\end{algorithm}

We provide the following theorem for pure differential privacy of the oneshot Laplace mechanism. Its proof uses a standard coupling argument and it is given in the supplementary material.

\begin{theorem}\label{thm:simple}
The oneshot Laplace mechanism is $(\e,0)$-differentially private if we set $\lambda = 2ks_f/\e$ or larger.
\end{theorem}

However, it is surprisingly challenging to prove the privacy guarantees for the oneshot Laplace mechanism in the approximate differential privacy framework. Here we state the theorem and leave the technical sketches and intuition to Section~\ref{sec:proof-archtecture} and the complete proof to the supplementary material.

\begin{theorem}[Privacy guarantees]
\label{thm:main}
Given $\epsilon \le 0.2$, $\delta \le 0.05$ and $m \ge 2$, the oneshot Laplace mechanism is $(\e,\delta)$-differentially private if we set $
\lambda_{\textnormal{oneshot}} = \frac{8s_f\sqrt{k\log(m/\delta)}}{\e}
$ or larger.
\end{theorem} 

We remark that $\epsilon$ can be set up to $O(\log(m/\delta))$ \citep{dwork2015private}, and the constant in $\lambda_{\textnormal{oneshot}}$ is for ease of analysis. We also point out that our result includes a higher multiplicative factor of $O(\sqrt{\log(m/\delta)})$ compared to $O(\sqrt{\log(1/\delta)})$ in the other results, which only incurs a small constant factor since $\delta$ is typically required to be $o(1/m)$. The next result is concerned with the utility of the oneshot Laplace mechanism. 

\begin{theorem}[Utility]
\label{thm:utility}
Let $f_{(1)}(D) \le f_{(2)}(D) \le \cdots \le f_{(m)}(D)$ denote the order statistics of the counts. Write $\Delta := \min_{1 \le i \le m-1} \left\{f_{(i+1)}(D) -  f_{(i)}(D)\right\}$.
Then with probability at least
\[
p(\Delta) = \max \left\{ 0, 1-\frac{(m-1)(2\lambda+\Delta)\ep{-\Delta/\lambda}}{4\lambda}  \right\},
\]
the oneshot Laplace mechanism returns the index set of the true top-$k$ elements.
\end{theorem}

The proof of Theorem~\ref{thm:utility} is mainly based on the application of Bonferroni bound and is left to the supplementary material. The form of $p(\Delta)$ guarantees that when the gaps of $f_i(D)$'s are significantly large, the oneshot Laplace mechanism can return the true index set of top-$k$ elements almost surely. Specifically, when $\Delta \ge 20 \lambda$ and $m \le 8\times 10^6$, then with probability at least $p(\Delta)>0.99$ the oneshot Laplace mechanism correctly returns the index set of the true top-$k$ elements.



\section{Application to Differentially Private Pairwise Comparison}
\label{sec:applications}

Our work was first motivated by and used for the private false discovery rate control mechanism~\citep{DworkDPFDRC18}. Here we present another application in ranking $n$ objects from partial binary comparisons, a problem with many important applications in Statistics and Computer Science.

Given a large collection of $m$ items, and only part of the comparisons $\{X_{ij} \}_{1 \le i \neq j \le m}$ between pairs of the $m$ items are revealed. Our goal is to privately recover the set of $k$ items with the highest ranks through the information released by pairwise comparison. One of the most widely used parametric models for pairwise comparison discovered is the \emph{Bradley-Terry-Luce (BTL) model} \citep{Bradley52,Luce59}. 

The BTL model was introduced to derive a full ranking when one only has access to pairwise comparison information. The basic idea of the Bradley-Terry-Luce parametric model is to assume that there exists a latent preference score $\omega_i^* (i=1, \cdots, m)$ assigned to the $m$ items of interest, and given a pair of items $(i,j)$ from the population, one can estimate the winning probability of item $j$ over item $i$ in the pairwise comparison as
\[
P_{ji}=\mathbb{P}\{ \text{item }j \text{ is preferred over item }i  \}= \frac{\omega_j^*}{\omega_i^*+\omega_j^*} \,.
\]

To define a comparison graph $\mathcal{G}=(\mathcal{V},E)$ for the Bradley-Terry-Luce model, we define the vertices set $\mathcal{V}=[m]$ of the graph $\mathcal{G}$ to represent the $m$ items we aim to compare, and each edge $(i,j)$ included in the edge set $E$ indicates that items $i$ and $j$ are compared and the comparison information is included in $\bm{y}$. We further assume that the comparison graph $\mathcal{G}$ is drawn from the Erd\"os-R\'enyi random graph \citep{erdos1960}, such that each edge between any two vertices is present independently with some probability that captures the fraction of paired items being compared. For each edge $(i,j) \in E$, we obtain $L$ independent paired comparison sampled from items $i$ and $j$, and for the $l$th comparison $y^{(l)}_{i,j}$, where $1\le l \le L$, we build the pairwise comparison model by assigning
\begin{equation}\label{eq:graphcomparison}
y^{(l)}_{i,j}=
\begin{cases}
1, & \text{with probability } \frac{\omega_j^\ast}{\omega_i^\ast+\omega_j^\ast},     \\
0, & \text{otherwise, }
\end{cases}
\end{equation}
and $y^{(l)}_{j,i}=1-y^{(l)}_{i,j}$ for all $(i,j) \in E$. The sufficient statistics of this model are given by
\[
\bm{y}:=\{y_{i,j} | (i,j) \in E \} \,,
\]
where 
$
y_{i,j}:=\frac{1}{L} \sum_{1 \le l \le L} y^{(l)}_{i,j} \,.$ We remark that in the regime of differential privacy, the comparison graphs of two adjacent datasets only differ in one data item, i.e., only one sample of a specific edge in the adjacent datasets is different.

Two algorithms tailored to the Bradley-Terry-Luce model that attract most attention are the \emph{spectral method (rank centrality)} and the \emph{maximum likelihood estimator method} \citep{spectral07}, the former of which we will focus on due to the applicability of the oneshot Laplace mechanism. To adopt the \emph{spectral method}, we make use of the pairwise comparison information $\bm{y}$ to establish a random walk over the graph $\mathcal{G}$ by defining its time-independent transition matrix $\bm{P}_{m \times m}=[P_{ij}]$, where $P_{ij}=\mathbb{P}(X_{t+1}=j|X_t=i)$ is defined as
\begin{equation}\label{eq:transition}
P_{ij}=
\begin{cases}
\frac{1}{d} y_{i,j}  & \text{if } (i,j) \in E,     \\
1-\frac{1}{d}\sum_{k:(i,k) \in E} y_{i,k} & \text{if } i=j, \\
0 & \text{otherwise.}
\end{cases} 
\end{equation}
Here $d>0$ is some given normalization factor which is on the same order of the maximum vertex degree of graph $\mathcal{G}$, and in general, we can assume that the normalization factor $d$ becomes larger as the number of vertices in graph $\mathcal{G}$ grows. The $\emph{spectral method}$ is summarized in Algorithm~\ref{algo:spectral}.

\begin{algorithm}[ht]
\caption{The spectral method for pairwise comparison}\label{algo:spectral}
\begin{algorithmic}[1]
\REQUIRE comparison graph $G=([m],E)$, sufficient statistics $\bm{y}$ and the normalization factor $d>0$
\ENSURE the rank of $\{ \pi(i)\}_{i \in [m]}$
\STATE define the defined transition matrix $\bm{P}$ as in (\ref{eq:transition})
\STATE compute the stationary distribution $\bm{\pi}$ of $\bm{P}$
\STATE sort $\pi_1,\cdots,\pi_m$ from low to high, $\pi_{(1)} \le \pi_{(2)} \le \cdots \le \pi_{(m)}$
\end{algorithmic}
\end{algorithm}

The intuition behind the $\emph{spectral method}$ is based on the fact that, assuming the sample size is sufficiently large, the stationary distribution $\bm{\pi}$ of the transition matrix $\bm{P}$ defined in (\ref{eq:transition}) is a reliable estimate of the preference scores $[\omega_1^\ast,\omega_2^\ast,\cdots,\omega_m^\ast]$ up to some scaling \citep{spectral07}. We notice that the result derived from the $\emph{spectral method}$ of pairwise comparison only takes advantage of the stationary distribution and, therefore, the oneshot Laplace mechanism can be applied to report top-$k$ elements via pairwise comparison information privately. We pause to introduce some definitions and a lemma to find the sensitivity of the statistic that maps the pairwise information to the stationary distribution. Throughout this paper, the $\infty$-norm $\|\bm{P}\|_{\infty}$ of a matrix $\bm{P}$ is its maximum absolute row sum. 

\begin{definition}[Ergodicity coefficient of a stochastic matrix~\citep{ergodicity2011}] 
\label{def:groupinverse}
For a $m \times m$ stochastic matrix $\bm{A}$, the ergodicity coefficient $\tau_1(\bm{A})$ of matrix $\bm{A}$ is defined as
\[
\tau_1(\bm{A}) \equiv \sup_{ \substack{\|\bm{v}\|_1=1 \\ \bm{v}^\top e=0}} \|\bm{v}^\top \bm{A}\|_1 \,,
\]
where $e$ is the vector of all ones.
\end{definition}

\begin{definition}[Conditional number of a Markov Chain~\citep{comparison2001}] 
\label{def:conditonalnum}
Let $\bm{P}$ denote the transition probability matrix of an $m$ state Markov chain $\mathcal{C}$, and $\bm{\pi}$ denotes the stationary distribution vector. The perturbed matrix $\tilde{\bm{P}}$ is the transition probability matrix of another $n$ state Markov chain $\tilde{\mathcal{C}}$ with stationary distribution vector $\tilde{\bm{\pi}}$. The conditional number $\kappa$ of a Markov chain $\mathcal{C}$ is defined by the following perturbation bound
$
\|\bm{\pi}-\tilde{\bm{\pi}}\|_{\infty} \le \kappa \|\bm{P}-\tilde{\bm{P}}\|_{\infty} \,.
$
\end{definition}

In \citet{ergodicity2011}, the authors stated the result that for every transition matrix $\bm{P}$, the ergodicity coefficient of $\bm{P}$ always falls between $0$ and $1$, and $\tau_1(\bm{P})=1$ if and only if the rank of matrix $\bm{P}$ equals 1. There is also a vast literature on exploring the form of the conditional number $\kappa$~\citep{comparison2001}. With all these preparations, we will build our private spectral method based on the following conclusion from \citet{seneta1988perturbation} and \citet{comparison2001}.

\begin{lemma}[Sensitivity of stationary distribution~\citep{seneta1988perturbation,comparison2001}]
\label{lm:stationarysensitivity}
Suppose $\bm{P}$ and $\tilde{\bm{P}}$ are $m \times m$ transition matrices with unique stationary distributions $\bm{\pi}^\top$ and $\tilde{\bm{\pi}}^\top$. If the ergodicity coefficient of transition matrix $\bm{P}$ satisfies $\tau_1(\bm{P})<1$, then
$
\| \bm{\pi}^\top - \tilde{\bm{\pi}}^\top\|_{\infty} \le \frac{1}{1-\tau_1(\bm{P})} \|\tilde{\bm{P}}-\bm{P}\|_{\infty}  \,.
$
\end{lemma}

Motivated by Lemma \ref{lm:stationarysensitivity}, the mapping $f$ has a bound sensitivity when the ergodicity coefficient of the transition matrix $\bm{P}$ is upper bounded by a constant $\rho<1$. In light of this observation, we can build our oneshot algorithm based on the following definition.

\begin{definition}[$\rho$-constrained comparison graph] 
\label{def:constrained}
A comparison graph $\mathcal{G}=(\mathcal{V},E)$ is said to be $\rho$-constrained if: (1) the transition matrix $\bm{P}$ of the Markov Chain defined as in (\ref{eq:transition}) has unique stationary distribution $\bm{\pi}^\top$. (2) there exists a constant $\rho<1$ such that $\tau_1(\bm{P}) \le \rho$.
\end{definition}

Definition \ref{def:constrained} implies that if the comparison graph $\mathcal{G}$ defined by the database $D$ is $\rho$-constrained, then the mapping $f: \bm{P} \rightarrow \bm{\pi}$ has a sensitivity bounded by $(1-\rho)^{-1}$. Making use of this fact, the oneshot Laplace mechanism for privately reporting the maximum $k$ elements through pairwise comparison information is stated in Algorithm~\ref{algo:spectraloneshot}.

\begin{algorithm}[ht]
\caption{The oneshot differentially private spectral method}\label{algo:spectraloneshot}
\begin{algorithmic}[1]
\REQUIRE $\rho$-constrained comparison graph $\mathcal{G}=([m],E)$, sufficient statistics $\bm{y}$, parameter $L$, normalization factor $d>0$, $k\geq 1$ and privacy parameters $\epsilon$, $\delta$
\ENSURE $i_1, \cdots, i_k$
\STATE define the transition matrix $\bm{P}$ as in (\ref{eq:transition})
\STATE compute the stationary distribution $\bm{\pi}=(\pi_1(\mathcal{G}),\cdots,\pi_m(\mathcal{G})))$ of $\bm{P}$
\STATE apply oneshot Laplace mechanism $\M^{os}$ to $-\pi_1(\mathcal{G}), \cdots, -\pi_m(\mathcal{G})$ with noise scale $\lambda$ to obtain $(i_1,y_{1}), (i_2, y_{2}), \cdots, (i_k, y_{k})$
\STATE return the set $\{i_1, \ldots, i_k\}$
\end{algorithmic}
\end{algorithm}

Now we establish the differential privacy of Algorithm~\ref{algo:spectraloneshot} in Theorem~\ref{thm:spectraloneshotprop} stated below, and the proof is left to the supplementary material.

\begin{theorem}\label{thm:spectraloneshotprop}
Given $\e \le 0.2$ and $\delta \le 0.05$, assume that the comparison graph $\mathcal{G}=([m],E)$ is $\rho$-constrained, then Algorithm~\ref{algo:spectraloneshot} is $(\e,\delta)$-differentially private if 
$
\lambda=\frac{8s\sqrt{k\log(m/\delta)}}{\e}
$ 
or larger, where the sensitivity $s=\frac{2}{dL(1-\rho)}$.
\end{theorem}



\section{Proofs and Intuition}
\label{sec:proof-archtecture}

In this section, we introduce a novel technique to prove the privacy of the oneshot Laplace mechanism. At a high level, the proof proceeds by considering the ``bad'' events, which have a large probability bias between two neighboring inputs. We show that those ``bad'' events happen when the sum of some dependent random variables deviates from its mean. We first partition the event space to remove the dependence between the random variables and, therefore, we can apply a concentration bound directly. Furthermore, we apply a coupling technique to pair up the partitions for the two neighboring inputs. For each pair, we apply a concentration inequality to bound the probability of ``bad'' events. The technical tools developed for proving the privacy of the oneshot top-$k$ algorithm in this section can be applied in many other settings. We provide the proof sketch to our main result and leave most of the technical details to the supplementary material.

Our goal is to reduce the dependence on~$k$ to~$\sqrt{k}$ for \emph{$(\epsilon, \delta)$}-differential privacy in the oneshot Laplace mechanism. We note that in the oneshot Laplace mechanism $\oneshot$, only a subset of $k$ elements, but not their ordering, is returned. The privacy proof of the oneshot Laplace mechanism crucially depends on this fact. We remark that one can further obtain the relative ranks and scores by running a second ranking phase in either mechanism to the reported $k$ elements. For example, by utilizing the Gaussian mechanism, one can publish more accurate scores, and hence their relative ranks, with the maximum noise of $O(\sqrt{k \log k})$ by paying slightly more privacy cost.

We start by providing the following lemma that directly establishes the privacy part of the oneshot Laplace mechanism in Theorem~\ref{thm:main}.

\begin{lemma}\label{lm:oneshot_private}
For any $k$-subset $S$ of $\{ 1, \cdots, m \}$ and any adjacent $x, x'$, we have 
\[
\P(\oneshot(x) \in S) \le \ep{\e}\P(\oneshot(x')\in S) + \delta\,.
\]
\end{lemma}

The key idea of the proof is to divide the event space by fixing the $k$th smallest noisy element $j$ together with the noise value $g_j$. For each partition, whether an element $i\neq j$ is selected by $\oneshot$ only depends on whether $x_i+g_i \leq x_j+g_j$, which happens with probability $q_i = G((x_j + g_j - x_i)/\lambda)$. Here $G$ denotes the cumulative distribution function of the standard Laplace distribution. As a result, we consider the following mechanism $\M$ instead: given $(q_1, \ldots, q_m)$ where $0<q_i<1$, output a subset of indices where each index $i$ is included in the subset with probability $q_i$. In the following proof, we will first understand the sensitivity of $q_i$ dependent on the change of $x_i$ and then show that $\M$ is ``private'' with respect to the corresponding sensitivity on $q$.
In order to prove Lemma~\ref{lm:oneshot_private}, we present the definition of $\tau$-closeness for vectors and two other lemmata we shall also use. 

\begin{definition}[$\tau$-closeness for vectors]
For two vectors $q=(q_1,\ldots, q_m)$ and $q'=(q'_1,\ldots,q'_m)$, we
say $q$ is \emph{$\tau$-close} with respect to $q'$ if for each $1\leq i\leq m$, $|q_i-q'_i|\leq
\tau q_i(1-q_i)$.
\end{definition}

\begin{lemma}\label{lem:close}
For any $z,z'$, we have 
\[
|G(z')-G(z)|\leq 2 \ep{|z'-z|}|z'-z| G(z)(1-G(z)), 
\] 
here $G$ denotes the cumulative distribution function of the standard Laplace distribution. 
\end{lemma}

The following lemma is the key step that constitutes the privacy guarantee of the mechanism $\M$ with respect to the sensitivity on $q$, and the proof of Lemma~\ref{lm:oneshot_private} is largely based on this result combined with Lemma~\ref{lem:close}.

\begin{lemma}\label{lm:dp_q_new}
Assume $C_0=3.9^2$, $C_1=1.95$. Under the conditions $\e \le 0.2$, $\delta \le 0.05$, $m \ge2$ and $k\geq C_0\log(m/\delta)$, and if $q$ is $\tau$-close with respect to $q'$ with $\tau \leq \frac{\e}{C_1\sqrt{k\log(m/\delta)}}$, then for any set $S$ of
$k$-subsets of $\{ 1, \cdots, m \}$, we have 
\[
\P(\M(q)\in S) \le \ep{\e}\P(\M(q')\in S) + \delta/m \,.
\]
\end{lemma}

The proof of Lemma~\ref{lem:close} is relatively straightforward and is relegated to the supplementary material. To prove Lemma~\ref{lm:dp_q_new}, notice that if $k\leq C_0\log(m/\delta)$, then according to Theorem~\ref{thm:simple}, the mechanism is $(\e,0)$-private. Thus we assume $k\geq C_0\log(m/\delta)$. Since $S$ consists of $k$-sets, we first show that if $\sum_i q_i \gg k$, then $\P(\M(q)\in S)$ is small. This can be done by applying the standard concentration bound in the following lemma. 

\begin{lemma}\label{lm:poisson_binomial}
Let $Z_1,\ldots, Z_m$ be $m$ independent Bernoulli random variables
with $\P(Z_i = 1) = q_i$. Suppose $\sum_{i=1}^m q_i \ge (1 + t)k$ for any $t > 0$. Then 
$
\P\left(\sum_i Z_i \leq k \right) \le \exp\left( -(1+t)kh\left(\frac{t}{t+1}\right)\right),
$ 
where $h(u) = (1+u)\log(1+u) - u$. Specifically, by setting $K = (1+c\sqrt{\log(m/\delta)/k})k$ and $c=1.9$, if we have $ \sum_{i=1}^m q_i \ge K$, then 
\[\P(\sum_{i=1}^{m} Z_i \leq k)\le \frac{\delta}{m}.\]
\end{lemma}

The proof of Lemma~\ref{lm:poisson_binomial} is based on the classical Bennett's inequality and is left to the supplementary material. By Lemma~\ref{lm:poisson_binomial}, we only need to consider the case of $\sum_i q_i \leq K$, which is more difficult than the case above. We first represent a set $S\subseteq\{1,\ldots,m\}$ by a binary vector $z\in \{0, 1\}^m$ and write $\P_{q}(z)$ as 
$
\P_q(z) = \prod_{i:z_i = 1}q_i\prod_{i:z_i=0}(1-q_i)\,.
$ 
Our goal is to show that for any $S$ consisting of weight $k$ vectors in $\{0,1\}^m$,
\[\sum_{z\in S} \P_q(z) \leq \ep{\e}\sum_{z\in S}\P_{q'}(z) + \frac{\delta}{m}.\]
By defining the set $S^\ast = \{z\;:\; \P_q(z) \geq \ep{\e}\P_{q'}(z)\}$, to prove Lemma~\ref{lm:dp_q_new}, it suffices to show that $\sum_{z\in S^\ast} \P_q(z) \leq \frac{\delta}{m}$. By the form of $\P_{q}(z)$, $z\in S^\ast$ holds if and only if
\begin{equation}\label{eq:diff}
\prod_{i:z_i = 1}q_i\prod_{i:z_i=0}(1-q_i) \geq \ep{\e}\prod_{i:z_i = 1}q'_i\prod_{i:z_i=0}(1-q'_i)\,.
\end{equation}
Let $\Delta_i = q'_i - q_i$. The $\tau$-closeness of $q$ with respect to $q'$ implies $|\Delta_i|\leq \tau q_i(1-q_i)$. Taking the logarithm of both sides of (\ref{eq:diff}) and rearranging gives
\[\sum_{i:z_i=1} \log(1+\Delta_i/q_i) + \sum_{i:z_i=0}\log(1 - \Delta_i/(1-q_i))\leq -\e\,.\]
To bound $\sum_{z\in S^\ast} \P_q(z)$, we consider independent Bernoulli random variables $Z_1, \ldots, Z_m$, where for each $i$, $Z_i = 1$
with probability $q_i$ and $Z_i=0$ with probability $1-q_i$. We set
$\zeta_i = Z_i\log(1+\Delta_i/q_i) + (1 - Z_i)\log(1 - \Delta_i/(1-q_i))$.
Note that 
\[
\sum_{z\in S^\ast} \P_q(z)=\sum_z  \mathbbm{1}_{\{ \sum \zeta_i \leq -\epsilon\}} \P_q(z) = \P\left(\sum_i \zeta_i \leq -\e\right),
\]
where $\mathbbm{1}_{\{\cdot\}}$ denotes the indicator function and the last probability is over the distribution of $Z_1,\ldots,Z_m$. Combine this with previous arguments, we need to prove that $\P(\sum_i \zeta_i \leq-\e)\leq \delta/m$.  It is easy to check that $\zeta_1 + \ldots +
\zeta_m$ has mean
$\sum_{i=1}^m \left( q_i\log(1+\Delta_i/q_i) + (1-q_i)\log(1-\Delta_i/(1-q_i))\right)$
and variance $\sigma^2 = \sum_{i=1}^m q_i(1-q_i)\log^2\frac{1+\Delta_i/q_i}{1-\Delta_i/(1-q_i)}$. To apply Bennett's inequality, we also need to check that the ranges of the centered random variables $\tilde{\zeta}_i := \zeta_i - q_i\log(1+\Delta_i/q_i) -(1-q_i)\log(1-\Delta_i/(1-q_i))$ are bounded in absolute value by $\max_{1 \le i \le m}\left|\log\frac{1 + \Delta_i/q_i}{1-\Delta_i/(1-q_i)}\right|$. To see why this is true, observe that 
\[
\begin{aligned}
\left|\tilde{\zeta}_i\right|=&\left|(Z_i - q_i) \log \frac{1+\Delta_i/q_i}{1-\Delta_i/(1-q_i)}\right|  \\
\leq{}&\max_{1 \le i \le m}\left|\log\frac{1 + \Delta_i/q_i}{1-\Delta_i/(1-q_i)}\right|.
\end{aligned}
\]
Therefore, according to Bennett's inequality we can assert that for any $t \ge 0$,
\[
\begin{aligned}
 \sum_{i=1}^m \zeta_i \ge \sum_{i=1}^m \left( q_i\log(1+\frac{\Delta_i}{q_i}) 
+ (1-q_i)\log(1-\frac{\Delta_i}{1-q_i})\right) - t \\
 \end{aligned}
\]
with probability at least $1 - \exp\left(- \frac{  \sigma^2h(At/\sigma^2)}{A^2} \right)$, where 
\[
A=\max_{1 \le i \le m}|\log\frac{1 + \Delta_i/q_i}{1-\Delta_i/(1-q_i)}|.
\]
 Furthermore, by taking $t = \e + \sum_{i=1}^m \left( q_i\log(1+\Delta_i/q_i) + (1-q_i)\log(1-\Delta_i/(1-q_i))\right)$, Bennett's inequality implies that $\P(\sum_i \zeta_i\leq -\e) \le \exp\left(-\sigma^2h(At/\sigma^2)/A^2 \right)$. Hence, the case of $\sum_i q_i \leq K$ can be established by proving
\begin{equation}\label{eq:exp_bound_delta}
\frac{\sigma^2h(At/\sigma^2)}{A^2}  \ge \log\frac{m}{\delta}\,.
\end{equation}

Now we seek to bound $t, \e, A$ and $\sigma$ using the fact that $|\Delta_i| \le \tau q_i(1-q_i)$. We start with exploring the relationship between $t$ and $\e$ by applying the standard results that
$
\log(1+u) \le u\,,
$
 and when $|u|\le \frac{1}{2}$,
$
\log(1+u) \geq u-u^2\,.
$
Notice that
\begin{multline}\nonumber
\max\left \{\left| \frac{\Delta_i}{q_i} \right|,\left| \frac{\Delta_i}{1-q_i}\right| \right \} \le \tau \le \frac{\epsilon}{ C_1 \sqrt{k\log(m/\delta)}} \\
\le \frac{\epsilon}{C_1 \sqrt{C_0} \log(m/\delta)} \le \frac{0.2}{1.95 \times 3.9 \times \log(2/0.05)} < \frac{1}{2} \,.
\end{multline}
We distinguish two cases. When $\Delta_i \geq 0$, we see that $q_i\log(1+\Delta_i/q_i)>0$ and $(1-q_i)\log(1-\Delta_i/(1-q_i))<0$. If $\left| q_i\log(1+\Delta_i/q_i) \right| \geq \left| (1-q_i)\log(1-\Delta_i/(1-q_i)) \right|$, these relations yield that
\[
\begin{aligned}
&\left| q_i\log(1+\Delta_i/q_i) + (1-q_i)\log(1-\Delta_i/(1-q_i))\right|\\
\leq{}&| q_i(\Delta_i/q_i+(\Delta_i/q_i)^2)+(1-q_i)(-\Delta_i/(1-q_i)\\
&+(\Delta_i/(1-q_i))^2) |\\
={}&q_i(\Delta_i/q_i)^2+(1-q_i)(\Delta_i/(1-q_i))^2\\
\leq{}&\tau^2\left(q_i(1-q_i)^2+q_i^2(1-q_i)\right)\\
\leq{}&\tau^2 q_i \,.\\
\end{aligned}
\]
Similarly, in the case that $\left| q_i\log(1+\Delta_i/q_i) \right| < \left| (1-q_i)\log(1-\Delta_i/(1-q_i)) \right|$, it follows that
\[
\begin{aligned}
&\left| q_i\log(1+\Delta_i/q_i) + (1-q_i)\log(1-\Delta_i/(1-q_i))\right|\\
\leq{}&| q_i(\Delta_i/q_i-(\Delta_i/q_i)^2)+(1-q_i)(-\Delta_i/(1-q_i)\\
&-(\Delta_i/(1-q_i))^2) |\\
={}&q_i(\Delta_i/q_i)^2+(1-q_i)(\Delta_i/(1-q_i))^2\\
\leq{}&\tau^2 \left(q_i(1-q_i)^2+q_i^2(1-q_i)\right)\\
\leq{}&\tau^2 q_i \,.\\
\end{aligned}
\]
To proceed, note that $\sum_i q_i\leq K\leq (1+\frac{c}{\sqrt{C_0}})k$, and thus
\[
\begin{aligned}
&\left|\sum_{i=1}^m \left( q_i\log(1+\Delta_i/q_i) + (1-q_i)\log(1-\Delta_i/(1-q_i))\right)\right|\\
\leq{}&\sum_{i=1}^m \left| q_i\log(1+\Delta_i/q_i) + (1-q_i)\log(1-\Delta_i/(1-q_i))\right|\\
\leq{}&\tau^2 \sum_{i=1}^m q_i \leq \left(1+\frac{c}{\sqrt{C_0}}\right)\tau^2 k \,.\\
\end{aligned}
\]
When $\Delta_i < 0$, by using the same arguments, we can also obtain
\[
\begin{aligned}
&\left|\sum_{i=1}^m \left( q_i\log(1+\Delta_i/q_i) + (1-q_i)\log(1-\Delta_i/(1-q_i))\right)\right| \\
\le{}&\left(1+\frac{c}{\sqrt{C_0}}\right)\tau^2 k \,.
\end{aligned}
\]
Making use of the assumption $\tau \le \e/(C_1\sqrt{k\log(m/\delta)})$, we observe
\[
\begin{aligned}
&\left| t-\e \right| \\
={}&\left|\sum_{i=1}^m \left( q_i\log(1+\frac{\Delta_i}{q_i}) + (1-q_i)\log(1-\frac{\Delta_i}{1-q_i})\right)\right|\\
\leq{}& \frac{1+\frac{c}{\sqrt{C_0}}}{C_1^2} \frac{\e^2}{\log(m/\delta)} \le \frac{1+\frac{1.9}{3.9}}{1.95^2}  \times \frac{0.2 \e}{\log(2/0.05)}  \\
\leq{}&0.0213 \e   \,.\\
\end{aligned}
\]
Rearranging the inequality above gives
\[
0.9787 \le \frac{t}{\e} \le 1.0213 \,.
\]
Furthermore, note that
\[
\begin{aligned}
\tau \le& \frac{\e}{C_1\sqrt{k\log(m/\delta)}} \le \frac{\epsilon}{C_1 \sqrt{C_0} \log(m/\delta)} \\
\le{}& \frac{0.2}{1.95 \times 3.9 \times \log(2/0.05)} <0.0072 \,.
\end{aligned}
\]
Hence,
\[
\begin{aligned}
&\left|\log\frac{1+\Delta_i/q_i}{1-\Delta_i/(1-q_i)}\right|\leq \left|\log\frac{1+\tau(1-q_i)}{1-\tau q_i}\right| \\
\leq{}& \frac{\tau}{1-\tau q_i}  \le \frac{\tau}{1-0.0072} < 1.0073\tau \,,
\end{aligned}
\]
which implies
\begin{equation}\label{eq:A}
A \le 1.0073 \tau \,.
\end{equation}
Combining the relations above implies that
\begin{align}
\sigma^2 &= \sum_{i=1}^m q_i(1-q_i)\log^2\frac{1+\Delta_i/q_i}{1-\Delta_i/(1-q_i)}\nonumber\\
&\leq 1.0073^2 \sum_{i=1}^m q_i(1-q_i)\tau^2\nonumber\\
&\leq 1.509  \tau^2 k\,.\label{eq:sigma}
\end{align}
Since $uh(a/u)$ is a decreasing function in $u$, from \eqref{eq:sigma} it follows that
\[
\sigma^2h(At/\sigma^2)\geq 1.509\tau^2k h(At/(1.509\tau^2k))\,,
\]
we set $\tau=\e/(C_1\sqrt{k\log(m/\delta)})$. Recognizing $k\geq C_0\log(m/\delta)$, it is clear that 
\[
\begin{aligned}
&\frac{At}{1.509\tau^2 k} \leq \frac{1.0073}{1.509} \cdot \frac{t}{\tau k} = \frac{1.0073 \times 1.95}{1.509}   \frac{t}{\e}  \sqrt{\frac{\log(m/\delta)}{k}} \\
\le{}& \frac{1.0073 \times 1.95}{1.509} \times 1.0213 \times \frac{1}{3.9} \le 0.3409 \,.
\end{aligned}
\]
Finally, by taking advantage of the fact that $h(u)/u^2$ is a decreasing function in $u$, we see that
\[
\begin{aligned}
&\frac{\sigma^2h(At/\sigma^2)}{A^2}\\
\geq{}&\frac{1.509 \tau^2k h(At/(1.509\tau^2k))}{A^2} \\
\geq{}& \frac{1.509\tau^2 k}{A^2} \cdot \frac{h(0.3409)}{0.3409^2} \cdot \left(\frac{At}{1.509\tau^2 k}\right)^2   \\
\geq{}& 1.509 \cdot \frac{h(0.3409)}{0.3409^2} \cdot \frac{1}{1.509^2} \cdot 1.95^2 \cdot \frac{ t^2 \log(m/\delta)}{\e^2}  \\
\geq{}& 1.509 \cdot \frac{h(0.3409)}{0.3409^2} \cdot \frac{1}{1.509^2} \cdot 1.95^2 \cdot (0.9787)^2 \log\left(\frac{m}{\delta}\right) \\
\geq{}& 1.08\log\left(\frac{m}{\delta}\right) \\
>{}& \log\left(\frac{m}{\delta}\right) \,.\\
\end{aligned}
\]
This completes the proof of inequality (\ref{eq:exp_bound_delta}), and therefore, also completes the proof of Lemma~\ref{lm:dp_q_new}. We now prove Lemma~\ref{lm:oneshot_private}. The proof follows from Lemma~\ref{lem:close} combined with Lemma~\ref{lm:dp_q_new}.

\begin{proof}[Proof of Lemma~\ref{lm:oneshot_private}]
Without loss of generality, we assume $s_f=1$. First, notice that if $k < C_0\log(m/\delta)$, 
\[
\lambda > \frac{8 \sqrt{k\log(m/\delta)}}{\e} \ge \frac{8}{\sqrt{C_0}} \cdot \frac{k}{\e} > \frac{2k}{\e} \,.
\]
Therefore, Theorem~\ref{thm:simple} immediately implies the mechanism is $(\e,0)$-private. Now we assume $k\geq C_0\log(m/\delta)$. Throughout we use $J_k$ to denote the random variable of the index of the $k$th smallest element in terms of the noisy count $x$, and we use $g_j$ to denote the noise added to the $j$th element of $x$. We also define $J'_k$ and $g'_j$ in terms of $x'$, respectively. For any given $J_k=j$ and the noise $g_j = g$, we have 
\[
\P(i\in\oneshot(x)) = G((x_j + g - x_i)/\lambda):=q_i.
\] 
Setting $q = q(g) = (q_i)$ for $1\leq i\leq m$ and $i\neq j$, and $S_j=\{s/\{j\}\;:\;s\in S\mbox{ and }j\in s\}$, then we have
\[
\P(\oneshot(x)\in S, J_k = j | g_j = g) = \P(\M(q)\in S_j).  
\]
Making use of the fact that $\|x-x'\|_{\infty}\leq 1$, we conclude that for any $i$, 
\[
\left|\frac{x_j+ g -x_i}{\lambda}-\frac{x'_j+ g -x'_i}{\lambda} \right|\leq \frac{2}{\lambda} \,.
\]
By Lemma~\ref{lem:close}, this implies
\[
\begin{aligned}
&\left|q-q' \right|\\
\le{}&2q(1-q)\ep{|\frac{x_j+ g -x_i}{\lambda}-\frac{x'_j+ g -x'_i}{\lambda}|} \cdot \\
&\>\>\left|\frac{x_j+ g -x_i}{\lambda}-\frac{x'_j+ g -x'_i}{\lambda} \right| \\
\le{}&2\ep{ \left|\frac{2}{\lambda} \right|} \left|\frac{2}{\lambda} \right| q(1-q) \le \frac{4}{\lambda}\ep{2\epsilon/  8\sqrt{k\log(m/\delta)}  }q(1-q) \\
\le{}&\frac{4}{\lambda}\ep{\epsilon/  4 \sqrt{C_0} \log(m/\delta)  }q(1-q)  \\
<{}&\frac{4.014}{\lambda}q(1-q) \,.
\end{aligned}
\]
Hence $q$ is  $4.014/\lambda$-close with respect to $q'$. We also notice that
\[
\lambda \geq \frac{8\sqrt{k\log(m/\delta)}}{\e} = \frac{C_1 \sqrt{k\log(m/\delta)}}{ 1.95\e/8}\,,
\]
which implies that $q$ is $\frac{0.9785\e}{C_1\sqrt{k\log(m/\delta)}}$-close with respect to $q'$. We write $\P(\oneshot(x)\in S, J_k = j | g_j = g)$ and $\P(\oneshot(x')\in S, J'_k = j | g'_j = g)$ as $\P_{x}$ and $\P_{x'}$ respectively. Applying Lemma~\ref{lm:dp_q_new} to $q$ and $q'$ with parameters $\e$ and
$\delta$, we have
\[
\P_{x}\leq \ep{0.9785\e} \P_{x'} + \delta/m \,.
\]
Let $l_{g_j}(g)$ stands for the probability when the $j$th noise is taking value of $g$. Noting that
\[
\lambda  \geq  \frac{8\sqrt{C_0}\log(m/\delta)}{\e} \geq  \frac{8 \times 3.9 \times \log(2/0.05)}{\e} > \frac{115}{\e} \,.
\]
The conclusion is now one step away. To show that the algorithm is $(\e,\delta)$-differentially private, note that
\[
\begin{aligned}
{}&\P(\oneshot(x)\in S)=\int \sum_{j=1}^{m} l_{g_j}(g)\P_{x} dg\\
\le{}&\int \sum_{j=1}^{m} l_{g_j}(g)\left[  \ep{0.9785\e} \P_{x'}+ \delta/m \right] dg\\
\le{}&\ep{0.9785\e}\int \sum_{j=1}^{m} \left( \frac{l_{g_j}(g)}{l_{g'_j}(g)} \right) \cdot l_{g'_j}(g) \cdot \P_{x'} dg+ \delta\\
={}&\ep{0.9785\e}\int \sum_{j=1}^{m} \left( e^{\frac{|g-x'_j|-|g-x_j|}{\lambda}} \right) \cdot l_{g'_j}(g)  \cdot \P_{x'} dg+ \delta \\
\le{}&\ep{0.9785\e+\frac{1}{\lambda}}\int \sum_{j=1}^{m}  l_{g'_j}(g)   \P_{x'} dg + \delta\\
\le{}&\ep{0.9785\e+\frac{\e}{115}}\int \sum_{j=1}^{m}  l_{g'_j}(g)   \P_{x'} dg+ \delta\\
<{}& \ep{0.99\e}\P(\oneshot(x')\in S)+\delta \,.
 \end{aligned}
\]
Therefore, Theorem~\ref{thm:main} is proved given the completion of the proof of Lemma~\ref{lm:oneshot_private}. \end{proof}



\section{Discussion}

In this paper, we provide a theoretical study of the classical top-$k$ problem in the regime of differential privacy. We propose a fast, low-distortion, statistically accurate, and differentially private algorithm to tackle this question, which we refer to as the oneshot Laplace mechanism. We provide a novel coupling technique in proving its privacy without taking advantage of the advanced composition theorems, thereby circumventing the linear dependence on $k$ in the privacy loss compared to the existing results in the literature. We further provide the applications of the oneshot Laplace mechanism in multiple hypothesis testing and pairwise comparison. Our contributions in the theoretical framework have the potential to impact many essential areas in machine learning in both theory and practice.

This study leaves a number of open questions that we hope will inspire further work. Through the proof of differential privacy on the oneshot Laplace mechanism, there is nothing to prevent us from achieving better bounds for $\lambda$. From a different angle, we wonder if the coupling technique would lead to tighter privacy analysis using other notions of privacy such as concentrated differential privacy, R{\'e}nyi differential privacy, and Gaussian differential privacy \citep{dwork2016concentrated,bun2016concentrated,mironov2017renyi,dong,bu2020deep}. Finally, an important direction for further research is to obtain sharp asymptotic properties of the oneshot mechanism and use the results to give a more comprehensive comparison between the oneshot Laplace mechanism and existing approaches in the literature.


\section*{Acknowledgments}

We are very grateful to Cynthia Dwork for encouragement and discussions on an early version of the manuscript. This work was supported
in part by NSF through CAREER DMS-1847415 and  CCF-1763314, a Facebook Faculty Research Award, and an Alfred
Sloan Research Fellowship.

\bibliographystyle{icml2021}
\bibliography{references}

\begin{thebibliography}{31}
\providecommand{\natexlab}[1]{#1}
\providecommand{\url}[1]{\texttt{#1}}
\expandafter\ifx\csname urlstyle\endcsname\relax
  \providecommand{\doi}[1]{doi: #1}\else
  \providecommand{\doi}{doi: \begingroup \urlstyle{rm}\Url}\fi

\bibitem[Abadi et~al.(2016)Abadi, Chu, Goodfellow, McMahan, Mironov, Talwar,
  and Zhang]{abadi2016deep}
Abadi, M., Chu, A., Goodfellow, I., McMahan, H.~B., Mironov, I., Talwar, K.,
  and Zhang, L.
\newblock Deep learning with differential privacy.
\newblock In \emph{Proceedings of the 2016 ACM SIGSAC Conference on Computer
  and Communications Security}, pp.\  308--318. ACM, 2016.

\bibitem[Bafna \& Ullman(2017)Bafna and Ullman]{bafna2017price}
Bafna, M. and Ullman, J.
\newblock The price of selection in differential privacy.
\newblock \emph{arXiv preprint arXiv:1702.02970}, 2017.

\bibitem[Banerjee et~al.(2012)Banerjee, Hegde, and
  Massouli{\'e}]{banerjee2012price}
Banerjee, S., Hegde, N., and Massouli{\'e}, L.
\newblock The price of privacy in untrusted recommendation engines.
\newblock In \emph{2012 50th Annual Allerton Conference on Communication,
  Control, and Computing (Allerton)}, pp.\  920--927. IEEE, 2012.

\bibitem[Bradley \& Terry(1952)Bradley and Terry]{Bradley52}
Bradley, R.~A. and Terry, M.~E.
\newblock Rank analysis of incomplete block designs: I. the method of paired
  comparisons.
\newblock \emph{Biometrika}, 39\penalty0 (3/4):\penalty0 324--345, 1952.

\bibitem[Bu et~al.(2020)Bu, Dong, Long, and Su]{bu2020deep}
Bu, Z., Dong, J., Long, Q., and Su, W.~J.
\newblock Deep learning with gaussian differential privacy.
\newblock \emph{Harvard Data Science Review}, 2020\penalty0 (23), 2020.

\bibitem[Bun \& Steinke(2016)Bun and Steinke]{bun2016concentrated}
Bun, M. and Steinke, T.
\newblock Concentrated differential privacy: Simplifications, extensions, and
  lower bounds.
\newblock In \emph{Theory of Cryptography Conference}, pp.\  635--658.
  Springer, 2016.

\bibitem[Chen et~al.(2017)Chen, Fan, Ma, and Wang]{spectral07}
Chen, Y., Fan, J., Ma, C., and Wang, K.
\newblock Spectral method and regularized {MLE} are both optimal for top-$ k $
  ranking.
\newblock \emph{arXiv preprint arXiv:1707.09971}, 2017.

\bibitem[Cho \& Meyer(2001)Cho and Meyer]{comparison2001}
Cho, G.~E. and Meyer, C.~D.
\newblock Comparison of perturbation bounds for the stationary distribution of
  a markov chain.
\newblock \emph{Linear Algebra and its Applications}, 335\penalty0
  (1-3):\penalty0 137--150, 2001.

\bibitem[Dong et~al.(2019)Dong, Durfee, and Rogers]{dong2019optimal}
Dong, J., Durfee, D., and Rogers, R.
\newblock Optimal differential privacy composition for exponential mechanisms
  and the cost of adaptivity.
\newblock \emph{arXiv preprint arXiv:1909.13830}, 2019.

\bibitem[Dong et~al.(2021)Dong, Roth, and Su]{dong}
Dong, J., Roth, A., and Su, W.~J.
\newblock Gaussian differential privacy.
\newblock \emph{Journal of the Royal Statistical Society, Series B}, 2021.
\newblock to appear.

\bibitem[Durfee \& Rogers(2019)Durfee and Rogers]{durfee2019practical}
Durfee, D. and Rogers, R.~M.
\newblock Practical differentially private top-k selection with
  pay-what-you-get composition.
\newblock In \emph{Advances in Neural Information Processing Systems}, pp.\
  3527--3537, 2019.

\bibitem[Dwork \& Roth(2014)Dwork and Roth]{dwork2014algorithmic}
Dwork, C. and Roth, A.
\newblock \emph{The Algorithmic Foundations of Differential Privacy}.
\newblock Foundations and Trends in Theoretical Computer Science. Now
  Publishers, 2014.

\bibitem[Dwork \& Rothblum(2016)Dwork and Rothblum]{dwork2016concentrated}
Dwork, C. and Rothblum, G.~N.
\newblock Concentrated differential privacy.
\newblock \emph{arXiv preprint arXiv:1603.01887}, 2016.

\bibitem[Dwork et~al.(2006{\natexlab{a}})Dwork, Kenthapadi, McSherry, Mironov,
  and Naor]{DworkKMMN06}
Dwork, C., Kenthapadi, K., McSherry, F., Mironov, I., and Naor, M.
\newblock Our data, ourselves: Privacy via distributed noise generation.
\newblock In \emph{Annual International Conference on the Theory and
  Applications of Cryptographic Techniques}, pp.\  486--503. Springer,
  2006{\natexlab{a}}.

\bibitem[Dwork et~al.(2006{\natexlab{b}})Dwork, McSherry, Nissim, and
  Smith]{DworkMNS06}
Dwork, C., McSherry, F., Nissim, K., and Smith, A.
\newblock Calibrating noise to sensitivity in private data analysis.
\newblock In \emph{Theory of cryptography conference}, pp.\  265--284.
  Springer, 2006{\natexlab{b}}.

\bibitem[Dwork et~al.(2010)Dwork, Rothblum, and Vadhan]{boosting}
Dwork, C., Rothblum, G.~N., and Vadhan, S.
\newblock Boosting and differential privacy.
\newblock In \emph{Foundations of Computer Science (FOCS), 2010 51st Annual
  IEEE Symposium on}, pp.\  51--60. IEEE, 2010.

\bibitem[Dwork et~al.(2015)Dwork, Su, and Zhang]{dwork2015private}
Dwork, C., Su, W., and Zhang, L.
\newblock Private false discovery rate control.
\newblock \emph{arXiv preprint arXiv:1511.03803}, 2015.

\bibitem[Dwork et~al.(2018)Dwork, Su, and Zhang]{DworkDPFDRC18}
Dwork, C., Su, W.~J., and Zhang, L.
\newblock Differentially private false discovery rate control.
\newblock \emph{arXiv preprint arXiv:1807.04209}, 2018.

\bibitem[Erdos \& R{\'e}nyi(1960)Erdos and R{\'e}nyi]{erdos1960}
Erdos, P. and R{\'e}nyi, A.
\newblock On the evolution of random graphs.
\newblock \emph{Publ. Math. Inst. Hung. Acad. Sci}, 5\penalty0 (1):\penalty0
  17--60, 1960.

\bibitem[Friedman \& Schuster(2010)Friedman and Schuster]{friedman2010data}
Friedman, A. and Schuster, A.
\newblock Data mining with differential privacy.
\newblock In \emph{Proceedings of the 16th ACM SIGKDD international conference
  on Knowledge discovery and data mining}, pp.\  493--502. ACM, 2010.

\bibitem[Hardt \& Roth(2013)Hardt and Roth]{hardt2013beyond}
Hardt, M. and Roth, A.
\newblock Beyond worst-case analysis in private singular vector computation.
\newblock In \emph{Proceedings of the forty-fifth annual ACM symposium on
  Theory of computing}, pp.\  331--340. ACM, 2013.

\bibitem[Ipsen \& Selee(2011)Ipsen and Selee]{ergodicity2011}
Ipsen, I.~C. and Selee, T.~M.
\newblock Ergodicity coefficients defined by vector norms.
\newblock \emph{SIAM Journal on Matrix Analysis and Applications}, 32\penalty0
  (1):\penalty0 153--200, 2011.

\bibitem[Kairouz et~al.(2017)Kairouz, Oh, and Viswanath]{KOV}
Kairouz, P., Oh, S., and Viswanath, P.
\newblock The composition theorem for differential privacy.
\newblock \emph{IEEE Transactions on Information Theory}, 63\penalty0
  (6):\penalty0 4037--4049, 2017.

\bibitem[Luce(2012)]{Luce59}
Luce, R.~D.
\newblock \emph{Individual choice behavior: {A} theoretical analysis}.
\newblock Courier Corporation, 2012.

\bibitem[McSherry \& Mironov(2009)McSherry and Mironov]{recom09}
McSherry, F. and Mironov, I.
\newblock Differentially private recommender systems: Building privacy into the
  netflix prize contenders.
\newblock In \emph{Proceedings of the 15th ACM SIGKDD international conference
  on Knowledge discovery and data mining}, pp.\  627--636. ACM, 2009.

\bibitem[McSherry \& Mironov(2013)McSherry and
  Mironov]{mcsherry2013differential}
McSherry, F.~D. and Mironov, I.
\newblock Differential privacy preserving recommendation, December~31 2013.
\newblock US Patent 8,619,984.

\bibitem[Mironov(2017)]{mironov2017renyi}
Mironov, I.
\newblock R{\'e}nyi differential privacy.
\newblock In \emph{2017 IEEE 30th Computer Security Foundations Symposium
  (CSF)}, pp.\  263--275. IEEE, 2017.

\bibitem[Qin et~al.(2016)Qin, Yang, Yu, Khalil, Xiao, and Ren]{qin2016heavy}
Qin, Z., Yang, Y., Yu, T., Khalil, I., Xiao, X., and Ren, K.
\newblock Heavy hitter estimation over set-valued data with local differential
  privacy.
\newblock In \emph{Proceedings of the 2016 ACM SIGSAC Conference on Computer
  and Communications Security}, pp.\  192--203. ACM, 2016.

\bibitem[Seneta(1988)]{seneta1988perturbation}
Seneta, E.
\newblock Perturbation of the stationary distribution measured by ergodicity
  coefficients.
\newblock \emph{Advances in Applied Probability}, 20\penalty0 (1):\penalty0
  228--230, 1988.

\bibitem[Shen \& Jin(2014)Shen and Jin]{shen2014privacy}
Shen, Y. and Jin, H.
\newblock Privacy-preserving personalized recommendation: An instance-based
  approach via differential privacy.
\newblock In \emph{2014 IEEE International Conference on Data Mining}, pp.\
  540--549. IEEE, 2014.

\bibitem[Steinke \& Ullman(2017)Steinke and Ullman]{steinke2017tight}
Steinke, T. and Ullman, J.
\newblock Tight lower bounds for differentially private selection.
\newblock In \emph{2017 IEEE 58th Annual Symposium on Foundations of Computer
  Science (FOCS)}, pp.\  552--563. IEEE, 2017.

\end{thebibliography}

\clearpage

\appendix

\section{Supplementary Material}

\subsection{Proof of Theorem~\ref{thm:simple}}

\begin{proof}[Proof of Theorem~\ref{thm:simple}] 
We prove this theorem with a standard coupling argument. Without loss of generality, we assume $s_f=1$. Let $x = (x_1, \ldots, x_m)$ and $x' = (x'_1, \ldots, x'_m)$ be adjacent, i.e., $\|x - x' \|_{\infty} \le 1$. Let $S$ be an arbitrary $k$-subset of $\{1, \ldots, m\}$ and $\mathcal{G}$ consist of all $(g_1, \ldots, g_m)$ such that $\oneshot$ with input $x$ reports $S$. Similarly we have $\mathcal{G}'$ for $x'$. It is clear that $\mathcal{G} - 2\cdot \mathbbm{1}_S \subset \mathcal{G}'$, where $\mathbbm{1}_S \in \{0, 1\}^m$ satisfies $\mathbbm{1}_S(i) = 1$ if and only if $i \in S$. Here $\{ 0,1 \}^m$ stands for the set of $m$-elements sets which only contain $0$ and $1$ as elements. Therefore, the standard coupling argument gives
\begin{multline}\nonumber
\prob(\oneshot(x) = S) = \int_{\mathcal{G}}\frac1{2^m\lambda^m}\ep{-\frac{\|g\|_1}{\lambda}} dg \\  
\ge \int_{\mathcal{G}' + 2\cdot\mathbbm{1}_S}\frac1{2^m\lambda^m}\ep{-\frac{\|g\|_1}{\lambda}} dg = \int_{\mathcal{G}'}\frac1{2^m\lambda^m}\ep{-\frac{\|g + 2\cdot\mathbbm{1}_S\|_1}{\lambda}} dg \\ 
\ge \int_{\mathcal{G}'}\frac{\ep{-2k/\lambda}}{2^m\lambda^m}\ep{-\frac{\|g\|_1}{\lambda}} dg  = \ep{-\e}\prob(\oneshot(x') = S)\,.
\end{multline}

On the opposite side, we have $\prob(\oneshot(x) = S) \le \ep{\e}\prob(\oneshot(x') = S)$, which completes the proof since $S$ is arbitrary.
\end{proof}

\subsection{Proof of Theorem~\ref{thm:utility}}
We begin by proving the following lemma.

\begin{lemma}\label{lm:difference}
Let $X$ and $Y$ be independent identically $\lap(\lambda)$ distributions and $Z=X-Y$, then the density function of random variable $Z$ has the form
\[
f_{Z}(z) = \frac{\lambda+|z|}{4\lambda^2} \cdot \exp\left(-\frac{|z|}{\lambda}\right).
\]
\end{lemma}

\begin{proof}[Proof of Lemma \ref{lm:difference}]
Notice that
\[
f_Z(z) = \int_{-\infty}^{\infty} f_X(x)f_Y(x-z)dx,
\]
by applying the convolution formula above, when $z \ge 0$ we have
\[
\begin{aligned}
&4\lambda^2f_Z(z)\\
={}& \int_{-\infty}^{\infty} \exp\left(-\frac{|x|}{\lambda}\right)\exp\left(-\frac{|x-z|}{\lambda}\right) dx \\
={}& \int_{-\infty}^{0} \exp\left(\frac{x}{\lambda}\right)\exp\left(\frac{x-z}{\lambda}\right) dx+ \\
&\int_{0}^{z} \exp\left(-\frac{x}{\lambda}\right)\exp\left(\frac{x-z}{\lambda}\right) dx+\\
&\int_{z}^{\infty} \exp\left(-\frac{x}{\lambda}\right)\exp\left(-\frac{x-z}{\lambda}\right) dx+\\
={}& (\lambda+z)\exp\left(-\frac{z}{\lambda}\right) \,.\\
\end{aligned}
\]
Therefore, with $z \ge 0$, we have 
\[
f_Z(z) = \frac{\lambda+z}{4\lambda^2}\exp\left(-\frac{z}{\lambda}\right).
\]
Since the pdfs of random variables $X$ and $Y$ are symmetric around the origin, the pdf of $Z$ must be symmetric around the origin. From this we get that
\[
f_Z(z) = \frac{\lambda+|z|}{4\lambda^2}\exp\left(-\frac{|z|}{\lambda}\right).
\]
\end{proof}
With the lemma above, we start to prove Theorem~\ref{thm:utility}. Let $g_1,g_2,\cdots,g_m$ be i.i.d. $\lap(\lambda)$ distributions. We define the event 
\[
A=\{\M_{oneshot}\text{ reports the index set of true top-}k \text{ elements}\}
\]
and consider the extreme case when $y_i$'s follow the exact same order as $f_i(D)$'s, i.e.,
\[
f_{(1)}(D)+g_1 \le f_{(2)}(D)+g_2 \le \cdots \le f_{(m)}(D)+g_m.
\]
Thus we can lower bound $P(A)$ by the extreme case above,
\[
\begin{aligned}
\P(A)\ge{}&\P(f_{(1)}(D)+g_1  \le \cdots \le f_{(m)}(D)+g_m)\\
={}& \P(f_{(1)}(D)+g_1  \le  f_{(2)}(D)+g_2,  \\
&f_{(2)}(D)+g_2 \le f_{(3)}(D)+g_3,\cdots, \\
&f_{(m-1)}(D)+g_{m-1} \le f_{(m)}(D)+g_m)\\
={}& \P(g_1-g_2  \le  f_{(2)}(D)-f_{(1)}(D),  \\
&g_2-g_3 \le f_{(3)}(D)-f_{(2)}(D),\cdots, \\
&g_{m-1}-g_m \le f_{(m)}(D)-f_{(m-1)}(D)) \\
\ge{}& \P(g_1-g_2  \le  \Delta, g_2-g_3 \le \Delta,\cdots, \\
&g_{m-1}-g_m \le \Delta) \,.\\
\end{aligned}
\]
Recall the Bonferroni lower bound that for events $X_1,\cdots,X_n$, we have
\[
\P(X_1,\cdots,X_n) \ge \max\left\{0,1-\sum_{i=1}^{m}\left(1-\P(X_i)\right)\right\}.
\]
Combining the calculation above, we have
\[
\begin{aligned}
\P(A)\ge{}&1-\sum_{i=1}^{m-1}\left(1-\P(g_i-g_{i+1} \le \Delta)\right) \\
={}& 1-\sum_{i=1}^{m-1}\P(g_i-g_{i+1} \le -\Delta)\\
={}& 1-\sum_{i=1}^{m-1}\int_{-\infty}^{-\Delta} f_Z(z)dz\\
={}& 1-\frac{(m-1)(2\lambda+\Delta)\ep{-\Delta/\lambda}}{4\lambda} \,,\\
\end{aligned}
\]
this completes the proof of Theorem~\ref{thm:utility}.

\subsection{Proof of Theorem~\ref{thm:spectraloneshotprop}}

\begin{proof}[Proof of Theorem~\ref{thm:spectraloneshotprop}] 
We only need to prove the sensitivity part $s=\frac{2}{dL(1-\rho)}$. For two adjacent databases $D$, $D'$ where they only differ in only one data item, we assume their corresponding sufficient statistics are $\bm{y}$ and $\tilde{\bm{y}}$. To capture the definition of adjacent databases, we assume one sample $y_{i_0,j_0}^{(l_0)}$ of the edge $(i_0,j_0)$ in database $D$ and $D'$ is different. Without loss of generality, we assume that $y_{i_0,j_0}^{(l_0)}=0$ in $D$ and $\tilde{y}_{i_0,j_0}^{(l_0)}=1$ in $D'$. Note that 
\[
P_{ij}=
\begin{cases}
\frac{1}{d} y_{i,j}  & \text{if } (i,j) \in E,     \\
1-\frac{1}{d}\sum_{k:(i,k) \in E} y_{i,k} & \text{if } i=j, \\
0 & \text{otherwise,}
\end{cases} 
\]
two transition matrices $\bm{P}$ and $\tilde{\bm{P}}$ only differ in four elements in positions $(i_0,j_0)$, $(j_0,i_0)$, $(i_0,i_0)$ and $(j_0,j_0)$. To find the maximum possible value of $||\bm{P}-\tilde{\bm{P}}||_{\infty}$, for $i \neq i_0$ and $i \neq j_0$,
\[
\sum_{j=1}^{m} |P_{ij}-\tilde{P}_{ij}|=0 \,.
\]
In the case when $i=i_0$,
\[
\begin{aligned}
&\sum_{j=1}^{m} |P_{ij}-\tilde{P}_{ij}|\\
={}& \frac{| y_{i_0,j_0} - \tilde{y}_{i_0,j_0} |}{d} + |P_{i_0i_0}-\tilde{P}_{i_0i_0}|\\
={}&\frac{| y_{i_0,j_0} - \tilde{y}_{i_0,j_0} |}{d} +\\
&\left| \left( 1-\frac{1}{d}\sum_{k:(i_0,k) \in E} y_{i_0,k}\right)-\left(1-\frac{1}{d}\sum_{k:(i_0,k) \in E} \tilde{y}_{i_0,k}\right)  \right|\\
={}&\frac{| y_{i_0,j_0} - \tilde{y}_{i_0,j_0} |}{d}+\frac{1}{d}\left| y_{i_0,j_0} - \tilde{y}_{i_0,j_0}  \right|\\
={}&\frac{2}{d}\left| y_{i_0,j_0} - \tilde{y}_{i_0,j_0} \right|  \\
={}&\frac{2}{dL}  \,.\\
\end{aligned}
\]
When $i=j_0$, we can follow the exact same calculation and get $\sum_{j=1}^{m} |P_{ij}-\tilde{P}_{ij}|=\frac{2}{dL}$. Therefore,
\[
||\bm{P}-\tilde{\bm{P}}||_{\infty}= \max_{1 \le i \le m} \sum_{j=1}^{m} |P_{ij}-\tilde{P}_{ij}| = \frac{2}{dL} \,.
\]
By the definition of sensitivity, we have $s=\frac{2}{dL(1-\rho)}$.
\end{proof}

\subsection{Proof of Lemma~\ref{lem:close}}

\begin{proof}[Proof of Lemma~\ref{lem:close}] 

By mean value theorem, there exists a point $\tilde{z}$ between $z$ and $z'$ such that
\[
\left|g(\tilde{z})\right|=\left|\frac{G(z')-G(z)}{z'-z}\right|\,,
\]
where $g$ denotes the density function of the standard Laplace distribution. Hence, we only need to prove
\[
\frac{\left|g(\tilde{z})\right|}{G(z)(1-G(z))} \le 2 \ep{|z'-z|}\,.
\]
Now we prove this inequality in different cases.
\begin{case}
when $\max(z,z') \le 0$. In this case,
\[
\frac{\left|g(\tilde{z})\right|}{G(z)(1-G(z))} = \frac{\ep{\tilde{z}-z}}{1-\frac{1}{2}\ep{z}} \le 2\ep{|\tilde{z}-z|} \le 2 \ep{|z'-z|}\,.
\]
\end{case}
\begin{case}
when $\min(z,z') \ge 0$. With a similar argument in Case 1, we have
\[
\frac{\left|g(\tilde{z})\right|}{G(z)(1-G(z))} = \frac{\ep{z-\tilde{z}}}{1-\frac{1}{2}\ep{-z}} \le 2\ep{|z-\tilde{z}|} \le 2 \ep{|z'-z|}\,.
\]
\end{case}
\begin{case}
when $\min(z,z') < 0 < \max(z,z')$. The triangle inequality gives
\[
\frac{\left|g(\tilde{z})\right|}{G(z)(1-G(z))} = \frac{\ep{|z|-|\tilde{z}|}}{1-\frac{1}{2}\ep{-|z|}} \le 2\ep{|z-\tilde{z}|} \le 2 \ep{|z'-z|}\,.
\]
\end{case}
\noindent In summary, combining all the cases above gives the lemma.
\end{proof}

\subsection{Proof of Lemma~\ref{lm:poisson_binomial}}

The proof of the Lemma~\ref{lm:poisson_binomial} is based on the classical Bennett's inequality stated below.

\textbf{Bennett's inequality}: Let $Z_1 \ldots, Z_n$ be independent random variables with all means being zero. In addition, assume $|Z_i| \le a$ almost surely for all $i$. Denoting by
$
\sigma^2 = \sum_{i=1}^n\mathrm{Var}(Z_i),
$
we have
\[
\P\left(\sum_{i=1}^n Z_i > t \right) \le \exp\left(-\frac{\sigma^2h(at/\sigma^2)}{a^2}\right)
\]
for any $t \ge 0$, where $h(u) = (1+u)\log(1+u) - u$.

\begin{proof}[Proof of Lemma \ref{lm:poisson_binomial}]
By the Bennett's inequality stated above, we have
\[
\begin{aligned}
&\P(\sum_i Z_i \leq k)\\
={}& \P\left(\sum_i Z_i - \sum_i q_i \leq -tk\right) \le \ep{-\sigma^2 h\left(tk/\sigma^2\right)}, 
\end{aligned}
\]
Note that $\P(\sum_i Z_i \leq k)$ is a decreasing function with respect to $\sum_{i=1}^{m} q_i$, hence we can assume $\sum_{i=1}^{m} q_i = (1+t)k$. In this case, we have $\sigma^2 = \sum_{i=1}^m q_i(1-q_i) \le (1+t)k$. Making use of the fact that $\sigma^2 h(tk/\sigma^2)$ is a monotonically decreasing function with respect to $\sigma^2$ gives
\[
\begin{aligned}
&\P(\sum_i Z_i \leq k)\le\ep{-\sigma^2 h\left(tk/\sigma^2 \right)}  \\
\le{}&  \exp\left( -(1+t)kh\left(\frac{t}{t+1}\right)\right)\,.
\end{aligned}
\]
Hence, the first part of Lemma~\ref{lm:poisson_binomial} is proved. In order to prove the second part of the lemma, we need to take advantage of the conclusion from the first part. Note that $h(u)/u^2$ is a decreasing function of $u$, by setting $t=c\sqrt{\frac{\log(m/\delta)}{k}} \le \frac{c}{\sqrt{C_0}}$, we get
\[
\begin{aligned}
{}&(1+t)kh\left(\frac{t}{t+1}\right)\\
\ge{}&(1+t)k\left(\frac{t}{t+1}\right)^2 \frac{h(\frac{c}{c+\sqrt{C_0}})}{(\frac{c}{c+\sqrt{C_0}})^2}\\
\ge{}&k\left(c\sqrt{\frac{\log(m/\delta)}{k}}\right)^2 \frac{1}{1+\frac{c}{\sqrt{C_0}}} \frac{h(\frac{c}{c+\sqrt{C_0}})}{(\frac{c}{c+\sqrt{C_0}})^2}\\
\ge{}&1.099 \log \left(\frac{m}{\delta} \right)\\
>{}&\log \left(\frac{m}{\delta} \right)  \,.\\
\end{aligned}
 \]
Therefore, when $\sum_{i=1}^m q_i \ge (1 + t)k$, we have
\[
\P(\sum_i Z_i \leq k) \le \exp\left( -(1+t)kh\left(\frac{t}{t+1}\right)\right) \le \frac{\delta}{m} \,.
\]
\end{proof}



\end{document}